\documentclass[sigconf,nonacm=True]{acmart}
\usepackage{mwe}
\usepackage{subfig}
\usepackage{multirow}

\AtBeginDocument{%
  \providecommand\BibTeX{{%
    \normalfont B\kern-0.5em{\scshape i\kern-0.25em b}\kern-0.8em\TeX}}}

\setcopyright{acmcopyright}
\copyrightyear{2020}
\acmYear{2020}
\acmDOI{10.1145/1122445.1122456}

\acmConference[MM '20]{Woodstock '18: ACM Conference}{October 12--16, 2020}{Seattle, United States}
\acmBooktitle{Woodstock '18: ACM Symposium on Neural Gaze Detection,
  June 03--05, 2018, Woodstock, NY}
\acmPrice{15.00}
\acmISBN{978-1-4503-XXXX-X/18/06}

\begin{document}

\title{Compare and Select: Video Summarization with Multi-Agent Reinforcement Learning}


\author{Tianyu Liu}
\affiliation{%
  \institution{Peking University}
  \city{Beijing}
  \country{China}}
\email{liutyb@gmail.com}

\renewcommand{\shortauthors}{Tianyu Liu.}

\begin{abstract}
  Video summarization aims at generating concise video summaries from the lengthy videos, to achieve better user watching experience. Due to the subjectivity, purely supervised methods for video summarization may bring the inherent errors from the annotations. To solve the subjectivity problem, we study the general user summarization process. General users usually watch the whole video, {\itshape compare} interesting clips and {\itshape select} some clips to form a final summary. Inspired by the general user behaviours, we formulate the summarization process as multiple sequential decision-making processes, and propose Comparison-Selection Network ({\itshape CoSNet}) based on multi-agent reinforcement learning. Each agent focuses on a video clip and constantly changes its focus during the iterations, and the final focus clips of all agents form the summary. The comparison network provides the agent with the visual feature from clips and the chronological feature from the past round, while the selection network of the agent makes decisions on the change of its focus clip. The specially designed unsupervised reward and supervised reward together contribute to the policy advancement, each containing local and global parts. Extensive experiments on two benchmark datasets show that {\itshape CoSNet} outperforms state-of-the-art unsupervised methods with the unsupervised reward and surpasses most supervised methods with the complete reward.
\end{abstract}




\maketitle

\section{Introduction}

Gigantic amounts of videos are produced by mobile phones, wearable devices and surveillance cameras. The lengthy raw videos with sparse information make it difficult for viewing, browsing and retrieving, resulting in the decline of user experience. In the meantime, video summaries can shorten the viewing time, provide dense information and save the storage space. To alleviate the problems of raw videos, we need video summarization to transform the lengthy raw videos into concise video summaries.

Video summarization is a relatively subjective task. In the process of creating datasets, different annotators may produce largely different annotations for the same video. Therefore, the annotation of video summarization datasets requires more annotators than other tasks, to ensure the maximum annotation accuracy. In the analysis of the widely used benchmark datasets, SumMe~\cite{gygli2014creating} and TVSum~\cite{song2015tvsum}, the former dataset has 15 to 18 annotations for each video, while the latter dataset has 20 annotations for each video. However, the annotations may still suffer from subjectivity due to the irreconcilable difference among different annotations.

\begin{figure}[t]
  \centering
  \includegraphics[width=\linewidth]{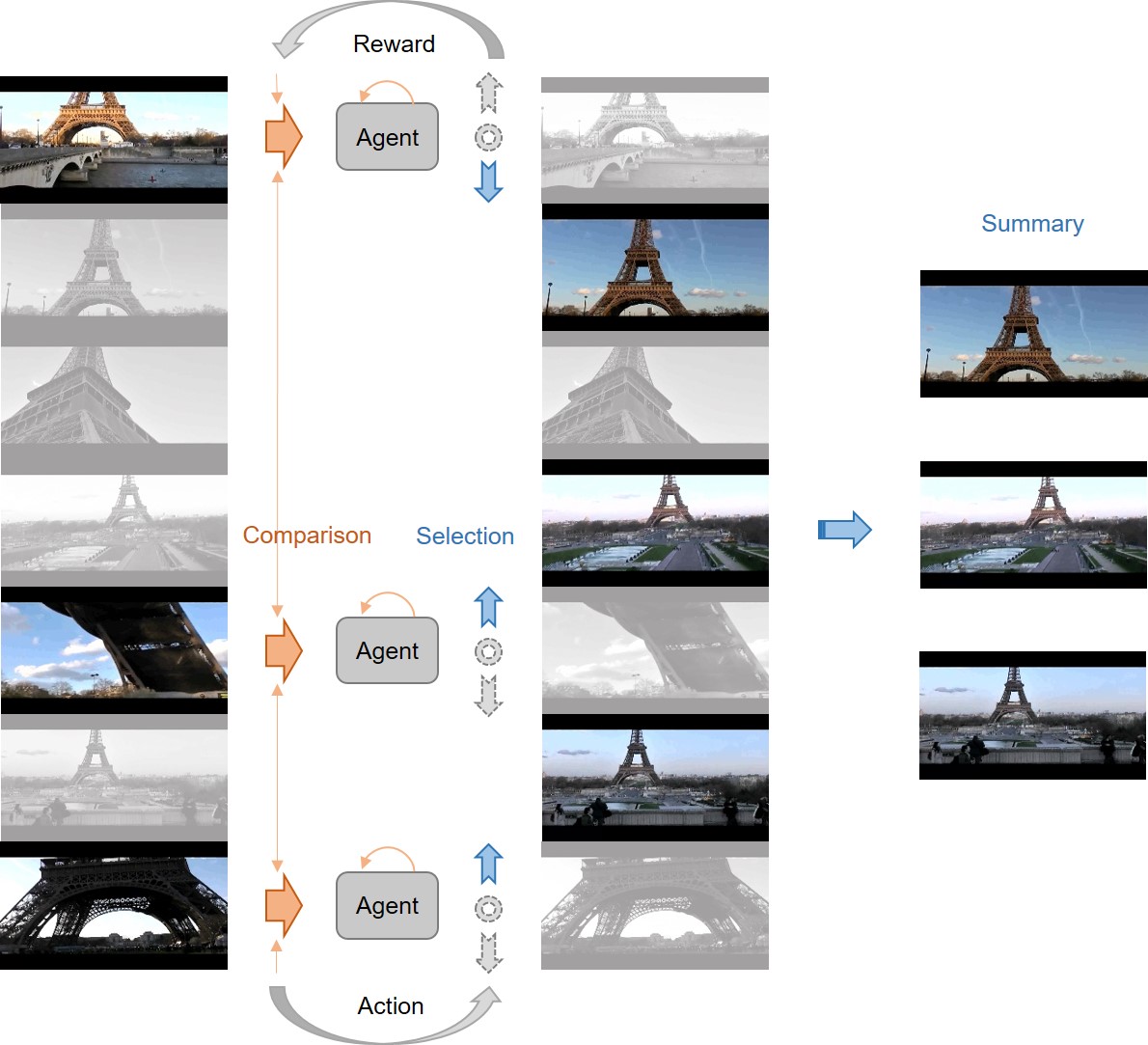}
  \caption{Agents {\itshape compare} the clips by taking the visual features and chronological features as input, then make decisions on which clips to {\itshape select} in the next round.}
  \Description{Agents {\itshape compare} the clips by taking the visual features and chronological features as input, then make decisions on which clips to {\itshape select} in the next round.}
  \label{fig-intro}
\end{figure}

In order to solve the subjectivity problem, we specially conduct a survey on the methods that can relieve the problem. Unsupervised methods~\cite{mahasseni2017unsupervised, zhang2018retrospective, he2019unsupervised, jung2019discriminative, yuan2019cycle} can work without annotations, but may lose some effective information in annotations. Among the various methods proposed in the literature, some methods originate from the inherent characteristics of the videos, and some other methods are from the inspiration of video user behaviours. In~\cite{song2015tvsum}, Song {\itshape et al.} proposed a method that selects frames which are most relevant to the video titles to form the summaries. Like video titles, Panda {\itshape et al.}~\cite{panda2017weakly} used video-level category annotations. The category annotations contain less information than frame-level annotations, but are actually more "accurate", because the opinions about the categories are consistent among almost all annotators. In~\cite{rochan2019video}, Rochan and Wang used the idea that unpaired videos and summaries can reduce the subjectivity caused by the interdependence of paired ones. In~\cite{xiong2019less}, Xiong {\itshape et al.} applied the thought of "less is more" to the video summarization task, which indicates that shorter videos are more informative than longer ones. The above-mentioned methods based on inherent characteristics or user behaviours can largely relieve the subjectivity problem.

We are also inspired by the behaviours of video users. How do general users summarize videos? They usually watch the whole video, {\itshape compare} the interesting clips ($p_1$ percent of the whole video length) and {\itshape select} the final clips to form the summary ($p_2$ percent, $p_2 < p_1 < 1.0$). We formulate this user summarization process as multiple sequential decision-making processes, and propose Comparison-Selection Network ({\itshape CoSNet}) that is composed of comparison network and selection network based on multi-agent reinforcement learning (MARL).

The $N$ agents of {\itshape CoSNet} "watch" $5N$ clips at the same time and change their focus clips constantly until all agents select the "stay-still" action or the iteration reaches the maximum, as shown in Fig.~\ref{fig-intro}. Each agent compares five clips (the focus clip, two neighboring clips and two focus clips of the neighboring agents) with the comparison network and selects the next clip to focus on with the selection network (by moving left, staying still or moving right). The comparison network is composed of 3D convolutional layers and long short-term memory (LSTM) layers, while the selection network contains fully connected (FC) layers. We specially design unsupervised reward and supervised reward, each with local and global parts. Unsupervised reward can avoid the subjectivity of the annotations, while supervised reward can bring some promotion from the annotation information. The local rewards act as the feedback for the local comparisons, while the global rewards are the feedback for the value of the global selection. In this way, {\itshape CoSNet} can follow the process that how general users summarize videos and relieve the subjectivity to the utmost extent.

To sum up, the major contributions of this paper are:
\begin{itemize}
  \item We follow the behaviours that general users usually summarize videos by comparing and selecting video clips, and formulate the user summarization process as multiple sequential decision-making processes.
  \item {\itshape CoSNet} is proposed based on MARL. It is composed of comparison network and selection network, with specially designed unsupervised reward and supervised reward (each with local and global parts).
  \item Extensive experiments on benchmark datasets show that {\itshape CoSNet} outperforms state-of-the-art unsupervised methods with the unsupervised reward and surpasses most supervised methods with the complete reward.
\end{itemize}

\section{Related Work}

\subsection{Video Summarization}

Some researchers formulate video summarization as an optimization problem~\cite{gygli2015video, xu2015gaze, zhang2016summary, elhamifar2017online, panda2017collaborative, mirzasoleiman2018streaming}. With the overwhelming trend of deep learning (DL), several kinds of DL based methods have also been applied to video summarization. Due to the temporal attributes of videos, some methods are based on different varieties of recurrent neural network (RNN)~\cite{zhang2016video, sigurdsson2016learning, zhao2017hierarchical, feng2018extractive, zhao2018hsa, wang2019stacked}, including LSTM~\cite{zhang2016video}, hierarchical RNN~\cite{zhao2017hierarchical, zhao2018hsa} and others. In~\cite{rochan2018video}, Rochan {\itshape et al.} formulated video summarization as a sequence labeling problem and used convolutional neural network (CNN) to solve it. Li {\itshape et al.}~\cite{li2018local} and Sharghi {\itshape et al.}~\cite{sharghi2018improving} proposed methods based on determinantal point processes (DPP). Methods based on unsupervised learning~\cite{mahasseni2017unsupervised, zhang2018retrospective, he2019unsupervised, jung2019discriminative, yuan2019cycle}, like generative adversarial networks (GAN) and variational autoencoders (VAE), try to make the summary features indistinguishable from the raw video features. Weakly supervised methods proposed by Cai {\itshape et al.}~\cite{cai2018weakly} and Panda {\itshape et al.}~\cite{panda2017weakly} are effective in that some videos have additional web information which can be used for summarization.

There are also some varieties of the video summarization task. The first variety is query-based video summarization~\cite{garcia2016first, sharghi2016query, vasudevan2017query, sharghi2017query, zhang2018query, choi2018contextually, xiao2020convolutional} that the summaries are generated according to user queries. The second variety is interactive video summarization~\cite{del2017active, jin2017elasticplay} that the computers interact with users during the summary generation processes. The third variety is 360-degree video summarization~\cite{lee2018memory, yu2018deep} that the 360-degree videos are summarized both temporally and spatially. The fourth variety is first-person video summarization~\cite{yao2016highlight, garcia2016first, poleg2015egosampling, silva2018weighted, ho2018summarizing, xu2015gaze, rathore2019generating} that the characteristics of the first-person videos are considered during the summarization process. In this paper, we focus on the general video summarization task.

\subsection{Reinforcement Learning}

The goal of reinforcement learning (RL) is to learn a good policy for the agent from experimental trials by maximizing expected future rewards. RL has made it to solve tasks in many research areas recently, such as games~\cite{mnih2015human, lample2017playing, ye2019mastering} and robotics~\cite{lillicrap2015continuous, gu2017deep, levine2016end}. It has succeeded in solving various vision tasks, like visual tracking~\cite{ren2018deep}, video face recognition~\cite{rao2017attention}, image cropping~\cite{li2018a2}, video captioning~\cite{pasunuru2017reinforced}, activity localization~\cite{wang2019language} and video classification~\cite{fan2018watching}. MARL also helps to solve some vision tasks. Rosello and Kochenderfer~\cite{rosello2018multi} proposed a method based on MARL for multi-object tracking. Wu {\itshape et al.}~\cite{wu2019multi} proposed a frame sampling method based on MARL for video recognition.

As the fast development, RL has also been applied to the video summarization task. In~\cite{zhou2018deep}, Zhou {\itshape et al.} proposed {\itshape DSN} with specially designed diversity and representativeness rewards. The rewards of {\itshape DSN} can describe how diverse and representative the generated summary is, but may ignore some local information. In~\cite{lan2018ffnet}, Lan {\itshape et al.} proposed {\itshape FFNet} for video fast-forwarding. {\itshape FFNet} is fast in processing speed, but many clips are omitted for "watching" that may bring some feature information loss. In~\cite{rathore2019generating}, Rathore {\itshape et al.} mainly focused on long egocentric video summarization. With MARL, our proposed {\itshape CoSNet} can simultaneously watch many clips to reduce feature information loss, and use both local and global rewards to reduce local information loss.

\section{Method}

\begin{figure*}[t]
  \centering
  \includegraphics[width=\textwidth]{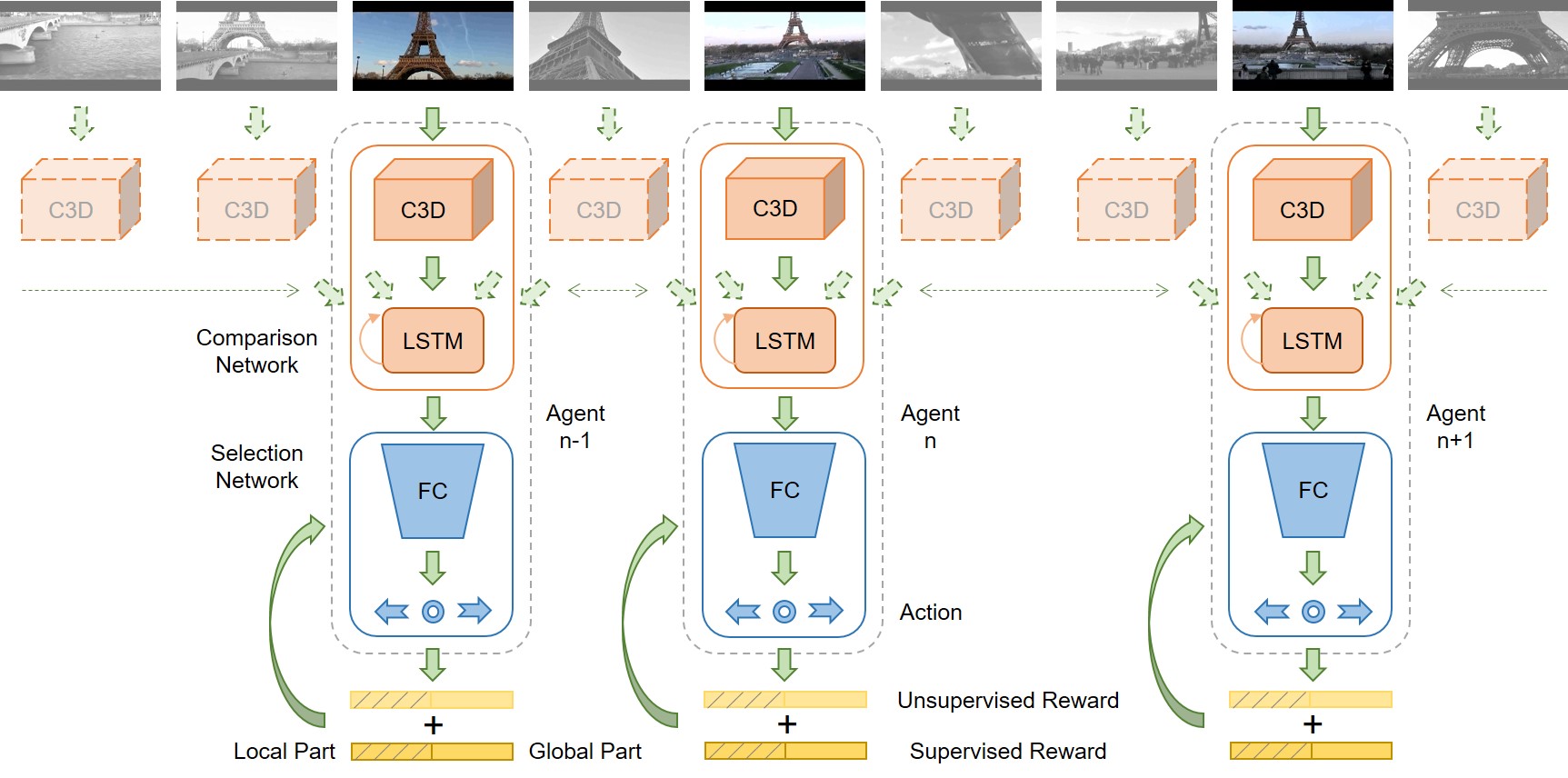}
  \caption{Each agent is composed of a comparison network and a selection network. Firstly, the C3D layers transform the raw clips into visual features. Then, the focus and neighboring features are averaged, as the input for the LSTM layers. Finally, the FC layers make the selection decisions based on the hidden states of the LSTM layers. The agents move left, stay still or move right according to the decisions. Unsupervised reward and supervised reward (each with local and global parts) are calculated during each round in the policy-based reinforcement learning processes.}
  \Description{Each agent is composed of a comparison network and a selection network. Firstly, the C3D layers transform the raw clips into visual features. Then, the focus and neighboring features are averaged, as the input for the LSTM layers. Finally, the FC layers make the selection decisions based on the hidden states of the LSTM layers. The agents move left, stay still or move right according to the decisions. Unsupervised reward and supervised reward (each with local and global parts) are calculated during each round in the policy-based reinforcement learning processes.}
  \label{fig-CoSNet}
\end{figure*}

We formulate video summarization as multiple sequential decision-making processes and propose {\itshape CoSNet} based on MARL. {\itshape CoSNet} contains $N$ agents. Each agent is composed of a comparison network and a selection network, with unsupervised reward and supervised reward (both with local and global parts). The agents are identical in network architecture and share the same parameters, for convenient experiments with different numbers of agents. Fig.~\ref{fig-CoSNet} is a demonstration of {\itshape CoSNet}.

\subsection{Problem Formulation}

The sequential decision-making process has four elements, i.e., state, action, policy and reward.

\paragraph{\textbf{State}}

The states $s \in S$ include the positions of the $N$ agents' focus clips, the visual features of clips and the chronological features. During each round, the agents have an temporal order and neighboring relations. We number the agents $\{a_i\}_{i=1}^N$ from left to right temporally by $1$ to $N$. The $M$ clips also have an temporal order and neighboring relations. We number the clips $\{c_j\}_{j=1}^M$ from left to right temporally by $1$ to $M$.

\paragraph{\textbf{Action}}

The actions $u \in U$ are discrete movements of the agents. An agent can move left, stay still or move right. The movement steps are within the scope of $u \in [-l, l]$. To avoid collisions, the agents execute the movement actions one by one. During the movement process, if an agent $a_i$ moves to a position that already has another agent $a_j$, $a_i$ moves further in the same direction until no collision exists. In particular, the video clips form a circle in our setting, which means the first clip $c_1$ is "adjacent" to the last clip $c_M$. Under this setting, an agent may move beyond $c_1$ and appear in one of the tail clips during left movements.

\paragraph{\textbf{Policy}}

The policy $\pi$ denotes the possibility of choosing action $u_t$ under state $s_t$, as Eq.~\eqref{eq:policy} shows.

\begin{equation}
  \pi(u_t|s_t) = \mathbb{P}_{\pi}[u=u_t|s=s_t]
  \label{eq:policy}
\end{equation}

\paragraph{\textbf{Reward}}

The reward $r$ of each round acts as important feedback for the advancement of the policy. The detailed reward definitions including unsupervised reward, supervised reward, local parts and global parts are presented in Sec.~\ref{sec:rd}.

\subsection{Network Architecture}
\label{sec:na}

\paragraph{\textbf{Comparison Network}}

The comparison network is composed of 3D convolutional layers for visual feature representation and LSTM layers for chronological feature reservation. In practice, the 3D convolutional layers are C3D~\cite{tran2015learning} layers pre-trained on Sports-1M dataset~\cite{karpathy2014large}. We use the "fc6" layer features $\{x_i\}_{i=1}^M$ of video clips (16 frames per clip).

Each agent $a_i$ focuses on a clip $c_j$ at round $t$. The input for the LSTM layers of $a_i$ is the average of the features from five clips (the focus clip of the left adjacent agent, the left adjacent clip, the focus clip, the right adjacent clip and the focus clip of the right adjacent agent). Then the LSTM layers produce hidden states $\{h_i\}_{i=1}^N$. For the first agent $a_1$, the left adjacent agent is the last agent $a_N$ under the circle setting. If agent $a_1$ focuses on the first clip $c_1$, the left adjacent clip of $a_1$ is the last clip $c_M$.

\paragraph{\textbf{Selection Network}}

The selection network contains two FC layers, to make decisions on which action to choose. The FC layers take the hidden states of the LSTM layers as input, output the action values for each possible action $u \in U$ and produce action choices $\{u_i\}_{i=1}^N$ with softmax.

\subsection{Reward Definition}
\label{sec:rd}

The reward definitions include unsupervised reward and supervised reward, each with both local and global parts.

\paragraph{\textbf{Unsupervised Reward}}

Unsupervised reward contains local unsupervised reward and global unsupervised reward, and does not need annotations for calculation.

Local unsupervised reward $r_t^{lu}$ (Eq.~\eqref{eq:r_lu}) denotes the local centrality of the focus clips. We want to improve the minimum feature similarity between the focus clip and its neighboring clips, to ensure that the focus clip approximates all its neighboring clips. The range of neighboring clips is between the focus clip of the left adjacent agent and the focus clip of the right adjacent agent. In Eq.~\eqref{eq:r_lu}, $c_{a_i}$ denotes the focus clip of agent $a_i$, $c'$ denotes a neighboring clip, $x(\cdot)$ denotes the C3D feature of the clip, and $\mathcal{N}(a_i)$ denotes the set of neighboring clips of agent $a_i$.

\begin{equation}
  r_t^{lu} = min_{c' \in \mathcal{N}(a_i)} (\frac{x(c_{a_i})^T x(c')}{\|x(c_{a_i})\|_2 \|x(c')\|_2})
  \label{eq:r_lu}
\end{equation}

Global unsupervised reward $r_t^{gu}$ (Eq.~\eqref{eq:r_gu}) denotes the overall feature difference among all focus clips, to ensure that the generated summaries cover most contents of the videos.

\begin{equation}
  r_t^{gu} = 1 - \frac{1}{N(N-1)} \sum_{i \in [1,N]} \sum_{j \in [1,N]}^{j \neq i} (\frac{x(c_{a_i})^T x(c_{a_j})}{\|x(c_{a_i})\|_2 \|x(c_{a_j})\|_2})
  \label{eq:r_gu}
\end{equation}

The unsupervised reward $r_t^u$ ($0 \leq r_t^u \leq 1$) is the combination of local unsupervised reward $r_t^{lu}$ ($0 \leq r_t^{lu} \leq 1$) and global unsupervised reward $r_t^{gu}$ ($0 \leq r_t^{gu} \leq 1$), as Eq.~\eqref{eq:r_u} shows. In Eq.~\eqref{eq:r_u}, $\alpha_1$ is a scale factor ($0 < \alpha_1 \leq 1$).

\begin{equation}
  r_t^u = \frac{1}{2} (r_t^{lu} + \alpha_1 r_t^{gu})
  \label{eq:r_u}
\end{equation}

\paragraph{\textbf{Supervised Reward}}

Supervised reward contains local supervised reward and global supervised reward, and needs annotations for calculation.

Local supervised reward $r_t^{ls}$ is composed of change reward ($[0,\frac{1}{2}]$, the first item in Eq.~\eqref{eq:r_ls}) and skip reward ($[0,\frac{1}{2}]$, the second item in Eq.~\eqref{eq:r_ls}). Change reward denotes the precision change between the new focus clip and the old focus clip, to ensure that the agents move towards more important clips with more key frames. When agents move, some frames are skipped over. Skip reward denotes the proportion of key frames in the frames skipped over, to ensure that the agents skip over less key frames. In Eq.~\eqref{eq:r_ls}, $f_t^{key}$ is the number of key frames in a clip at round $t$, $f^{clip}$ is the number of frames in a clip, $f_{\Delta t}^{key}$ is the number of key frames skipped over between $t-1$ and $t$, $f_{\Delta t}^{skip}$ is the number of frames skipped over between $t-1$ and $t$, $\alpha_2$ is a scale factor ($0 < \alpha_2 \leq 1$).

\begin{equation}
  r_t^{ls} = \frac{1}{4}(1 + \frac{f_t^{key}}{f^{clip}} - \frac{f_{t-1}^{key}}{f^{clip}}) + \frac{1}{2} \alpha_2 (1 - \frac{f_{\Delta t}^{key}}{f_{\Delta t}^{skip}})
  \label{eq:r_ls}
\end{equation}

Global supervised reward $r_t^{gs}$ (Eq.~\eqref{eq:r_gs}) denotes the global precision change between the new round and the old round, which also ensures that the agents move towards more important clips.

\begin{equation}
  r_t^{gs} = \frac{1}{2}(1 + \frac{\sum\nolimits_{agents} f_t^{key}}{\sum\nolimits_{agents} f^{clip}} - \frac{\sum\nolimits_{agents} f_{t-1}^{key}}{\sum\nolimits_{agents} f^{clip}})
  \label{eq:r_gs}
\end{equation}

The supervised reward $r_t^s$ ($0 \leq r_t^s \leq 1$) is the combination of local supervised reward $r_t^{ls}$ ($0 \leq r_t^{ls} \leq 1$) and global supervised reward $r_t^{gs}$ ($0 \leq r_t^{gs} \leq 1$), as Eq.~\eqref{eq:r_s} shows. In Eq.~\eqref{eq:r_s}, $\alpha_3$ is a scale factor ($0 < \alpha_3 \leq 1$).

\begin{equation}
  r_t^s = \frac{1}{2} (r_t^{ls} + \alpha_3 r_t^{gs})
  \label{eq:r_s}
\end{equation}

We can further combine the unsupervised reward and the supervised reward. The reward of agent $a$ in round $t$ is shown in Eq.~\eqref{eq:r_round}, in which $\alpha_4$ is a scale factor ($0 < \alpha_4 \leq 1$). $r_t^a$ also has the range of $0 \leq r_t^a \leq 1$.

\begin{equation}
  r_t^a = \frac{1}{2} (r_t^s + \alpha_4 r_t^u)
  \label{eq:r_round}
\end{equation}

The accumulated reward of agent $a$ from round $t$ on is in the form of Eq.~\eqref{eq:r_accumulated}. $\gamma$ ($0 < \gamma \leq 1$) is the discount factor.

\begin{equation}
  R_t^a = \sum\limits_{\tau=t}^T \gamma^{\tau-t}r_t^a
  \label{eq:r_accumulated}
\end{equation}

\subsection{Policy Gradient}

The objective function is defined as Eq.~\eqref{eq:objective}. Our optimization goal is to learn a policy $\pi$ with parameters $\theta_\pi$ by maximizing the objective function $J(\theta_\pi)$.

\begin{equation}
  J(\theta_\pi) = \sum\limits_{a=1}^N \sum\limits_{t=0}^T \sum\limits_{u \in U} \pi(u_t^a|s_t^a;\theta_\pi) R_t^a
  \label{eq:objective}
\end{equation}

The parameters $\theta_\pi = \{\theta_c, \theta_s\}$ need to be learned for optimization. $\theta_c$ represents the parameters of the LSTM layers, and $\theta_s$ represents the parameters of the FC layers, while the pre-trained parameters of C3D layers are frozen. The gradient of $J(\theta_\pi)$ is shown in Eq.~\eqref{eq:gradient}.

\begin{equation}
  \bigtriangledown_{\theta_\pi} J(\theta_\pi) = \sum\limits_{a=1}^N \sum\limits_{t=0}^T \sum\limits_{u \in U} \pi(u_t^a|s_t^a;\theta_\pi) \bigtriangledown_{\theta_\pi} log \pi(u_t^a|s_t^a;\theta_\pi) R_t^a
  \label{eq:gradient}
\end{equation}

The optimization is non-trivial due to the high-dimensional action sequences. Therefore, We follow REINFORCE~\cite{williams1992simple} algorithm to optimize the parameters. we approximate the policy gradient by K action sequences, as shown in Eq.~\eqref{eq:approximation}.

\begin{equation}
  \bigtriangledown_{\theta_\pi} J(\theta_\pi) \approx - \frac{1}{K} \sum\limits_{k=1}^K \sum\limits_{a=1}^N \sum\limits_{t=0}^T \bigtriangledown_{\theta_\pi} log \pi(u_{t,k}^a|s_{t,k}^a;\theta_\pi) R_t^a
  \label{eq:approximation}
\end{equation}

The loss function has two items, the policy gradient and a $l_2$ normalization term. We update the parameters $\theta_\pi$ by Eq.~\eqref{eq:loss}. $\eta$ is the learning rate, and $\lambda$ ($0 < \lambda \leq 1$) is a scale factor.

\begin{equation}
  \theta_\pi = \theta_\pi - \eta \bigtriangledown_{\theta_\pi} (J(\theta_\pi) + \lambda \sum_{\omega \in \theta_\pi} \omega^2)
  \label{eq:loss}
\end{equation}

\section{Experiments}

\subsection{Experimental Setup}
\label{sec:es}

\begin{figure*}[t]
  \centering
  \subfloat[][Sample frames of video-17 (Reuben Sandwich with Corned Beef $\&$ Sauerkraut) in TVSum.]{\includegraphics[width=\textwidth]{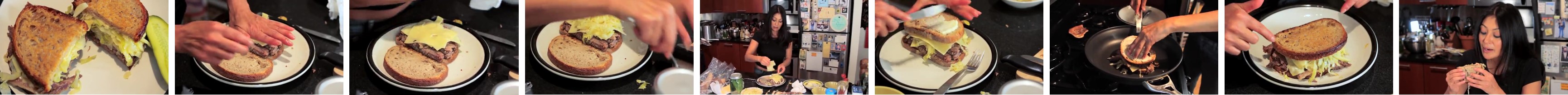}}\\
  \subfloat[][CoSNet-US.]{\includegraphics[width=.49\textwidth]{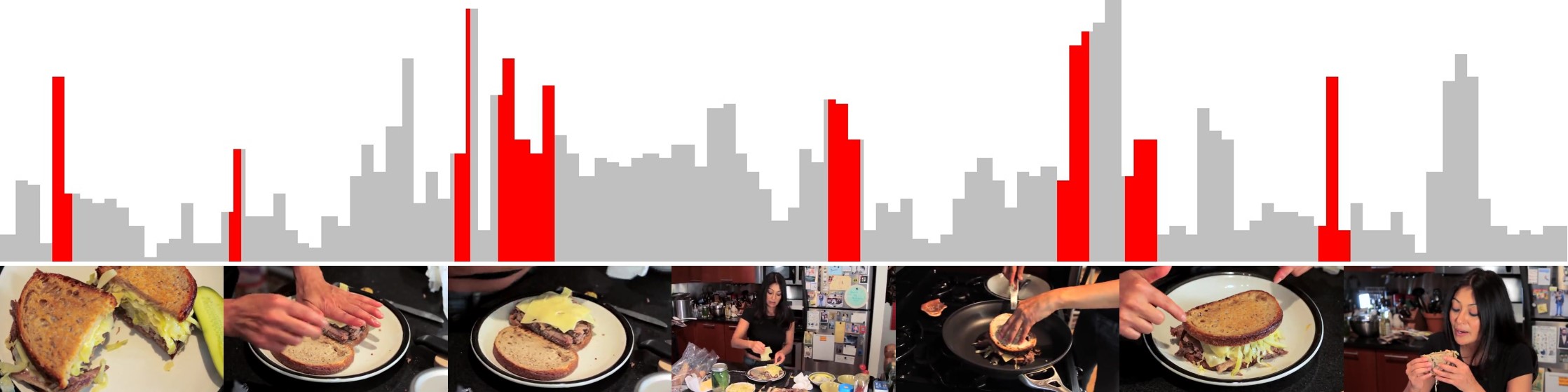}}\quad
  \subfloat[][CoSNet-U.]{\includegraphics[width=.49\textwidth]{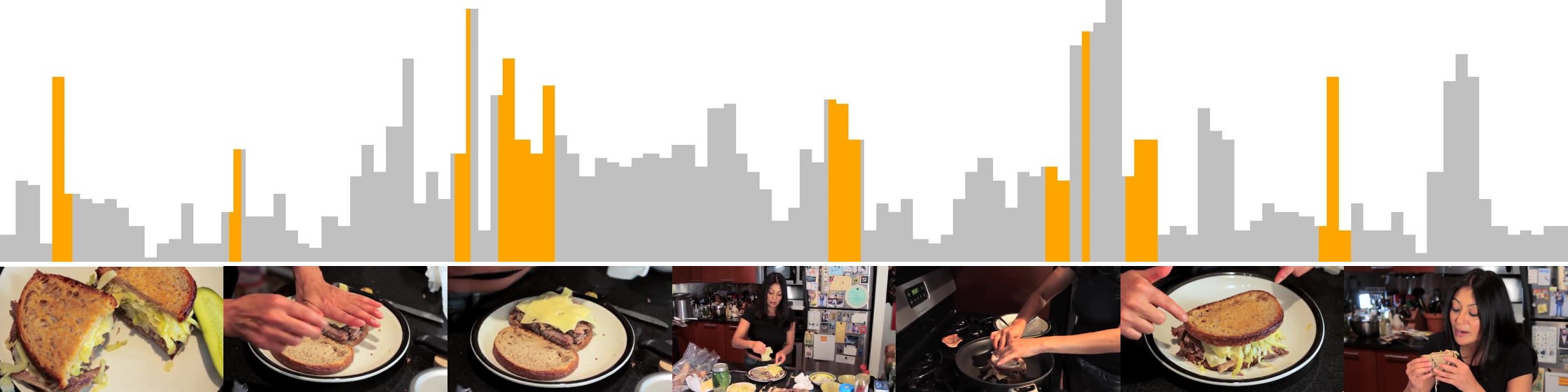}}\\
  \subfloat[][CoSNet-S.]{\includegraphics[width=.49\textwidth]{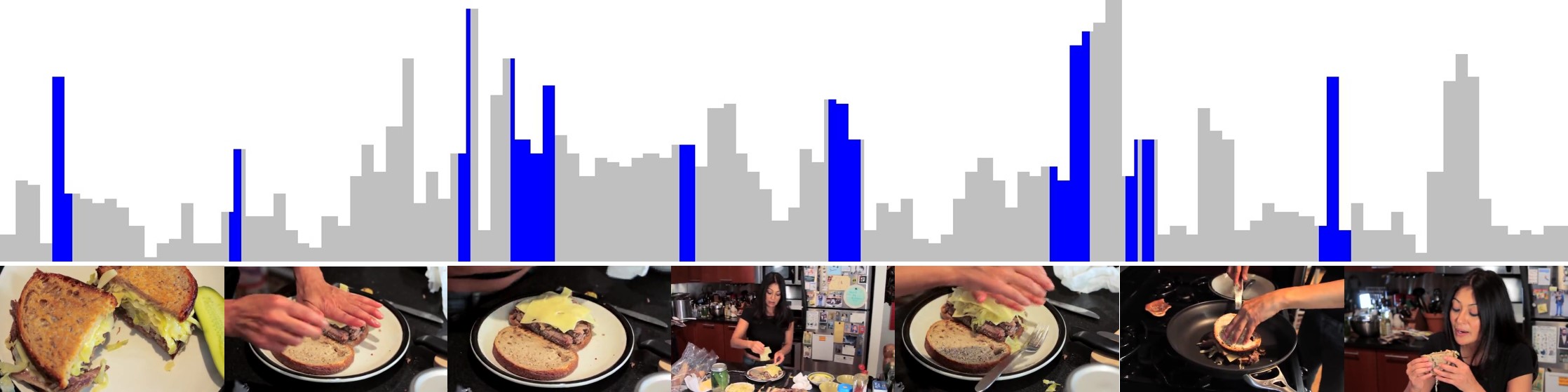}}\quad
  \subfloat[][CoSNet-LU.]{\includegraphics[width=.49\textwidth]{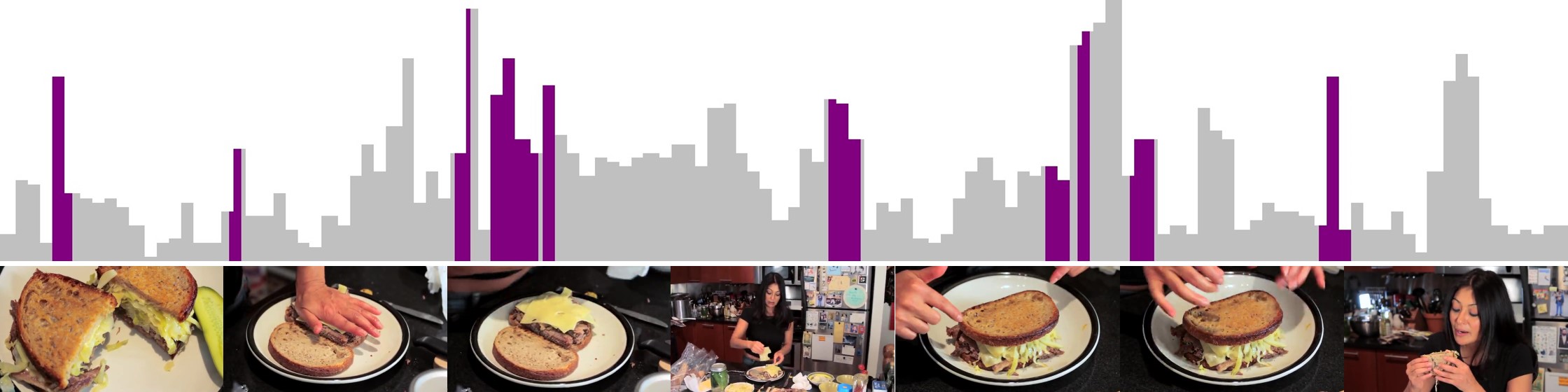}}
  \caption{(a) is the sample frames of video-17 in TVSum. The girl is cooking a special beef sandwich. The heights of all the bars in (b) to (e) represent the ground truth importance scores of video-17. The colored bars represent the selected segments with the corresponding methods.}
  \Description{(a) is the sample frames of video-17 in TVSum. The girl is cooking a special beef sandwich. The heights of all the bars in (b) to (e) represent the ground truth importance scores of video-17. The colored bars represent the selected segments with the corresponding methods.}
  \label{fig-one}
\end{figure*}

\paragraph{\textbf{Datasets}}

We conduct experiments on two widely used benchmark datasets, SumMe and TVSum. SumMe contains 25 videos of various topics, like sports and sights. The duration of the videos in SumMe varies from 1 to 6 minutes. TVSum contains 50 videos of 10 kinds. The duration of the videos in TVSum is between 2 and 10 minutes. The videos in both datasets are annotated by frame-level importance scores. The duration of a ground truth summary is around or below 15 percent of the raw video duration.

We also use the OVP~\cite{zhang2016video} dataset (50 videos) and the YouTube~\cite{de2011vsumm} dataset (39 videos) for data augmentation and transfer learning. In order to unify the different annotation forms of different datasets, we follow~\cite{zhang2016video} to make conversions among key-frames, key-shots and frame-level importance scores. During the above conversions processes, We use KTS~\cite{potapov2014category} to temporally divide the raw videos into segments.

\paragraph{\textbf{Evaluation Metric}}

Similar to other methods, we take the F-score based on key shots used in~\cite{zhang2016video} as the evaluation metric, to ensure fair comparisons. In the form of the harmonic mean of precision $P$ and recall $R$, F-score (Eq.~\eqref{eq:F-score}) depicts the similarity between the ground truth summaries and the generated summaries.

\begin{equation}
  F = \frac{2PR}{P+R} \times 100\%
  \label{eq:F-score}
\end{equation}

The experiments are conducted under the three kinds of settings in~\cite{zhang2016video}. Canonical (C) setting represents the standard 5-fold cross validation (5FCV) metric. Augmented (A) setting denotes augmenting the training set with other three datasets under 5FCV metric. Transfer (T) setting means only using other three datasets as the training set.

\begin{table}
  \caption{The evaluation results ($\%$) of different reward combinations on SumMe and TVSum.}
  \label{tab-reward}
  \begin{tabular}{c|c|c}
    \toprule
    Method & SumMe & TVSum \\
    \midrule
    CoSNet-GU & 43.5 & 56.9 \\
    CoSNet-GS & 44.6 & 57.3 \\
    CoSNet-LU & 45.9 & 58.4 \\
    CoSNet-LS & 46.3 & 58.7 \\
    CoSNet-U & 46.4 & 59.3 \\
    CoSNet-S & 47.1 & 59.4 \\
    \midrule
    \midrule
    CoSNet-US & 47.8 & 59.7 \\
    \bottomrule
  \end{tabular}
\end{table}

\paragraph{\textbf{Implementation Details}}

The 3D convolutional layers are pre-trained C3D layers as mentioned in Sec.~\ref{sec:na}. The initial positions of the $N$ agents are random and different among epochs. The $N$ agents are the same in network architecture and parameters during a round and the parameters update uniformly according to the experiences of all agents.

\begin{figure*}[t]
  \centering
  \subfloat[][video-18 (F-score: 72.1$\%$).]{\includegraphics[width=.49\textwidth]{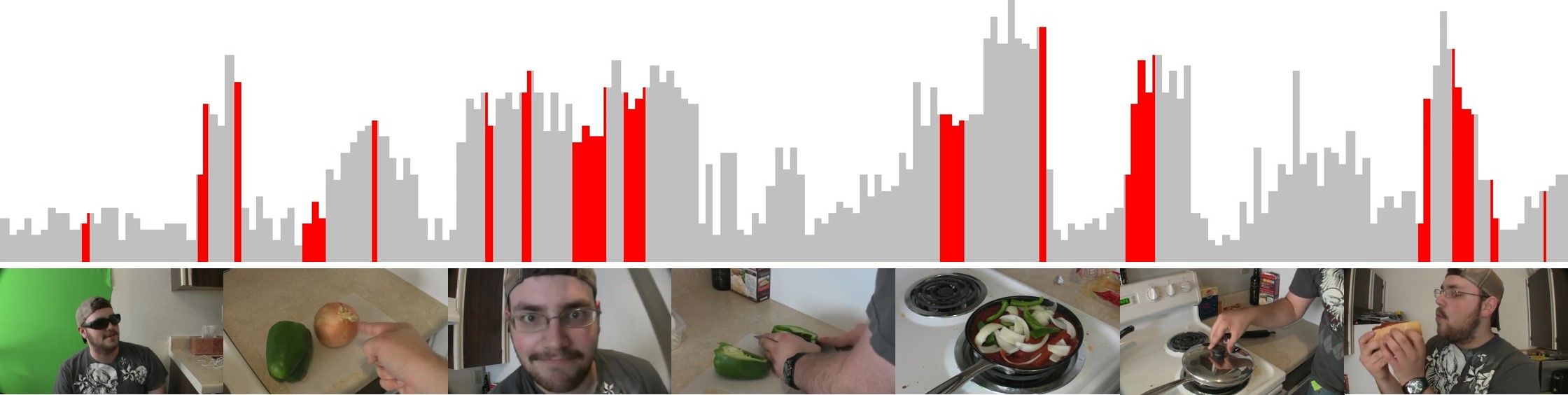}}\quad
  \subfloat[][video-22 (F-score: 59.6$\%$).]{\includegraphics[width=.49\textwidth]{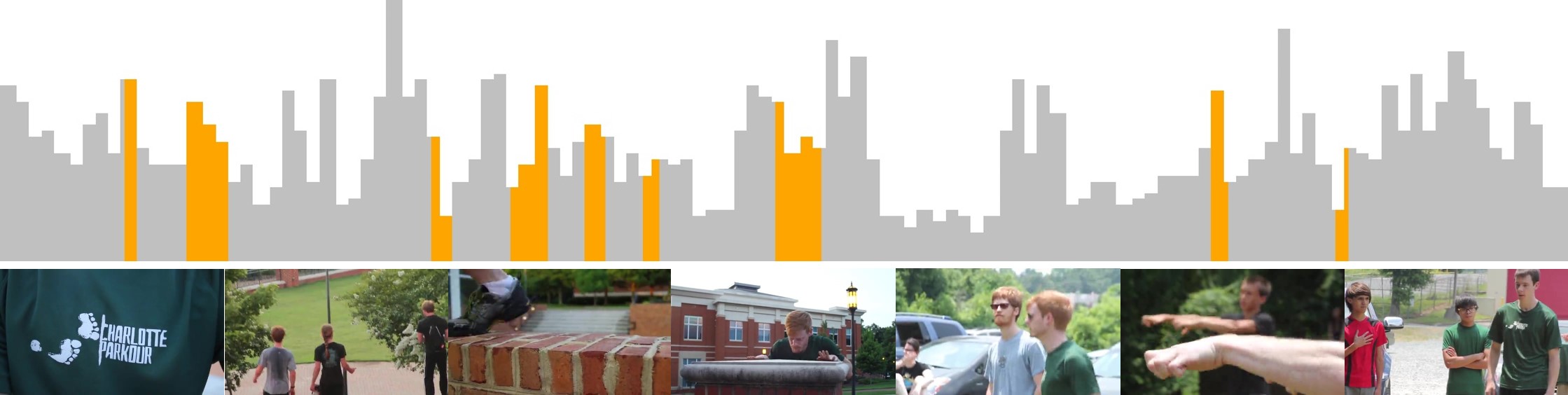}}\\
  \subfloat[][video-28 (F-score: 56.8$\%$).]{\includegraphics[width=.49\textwidth]{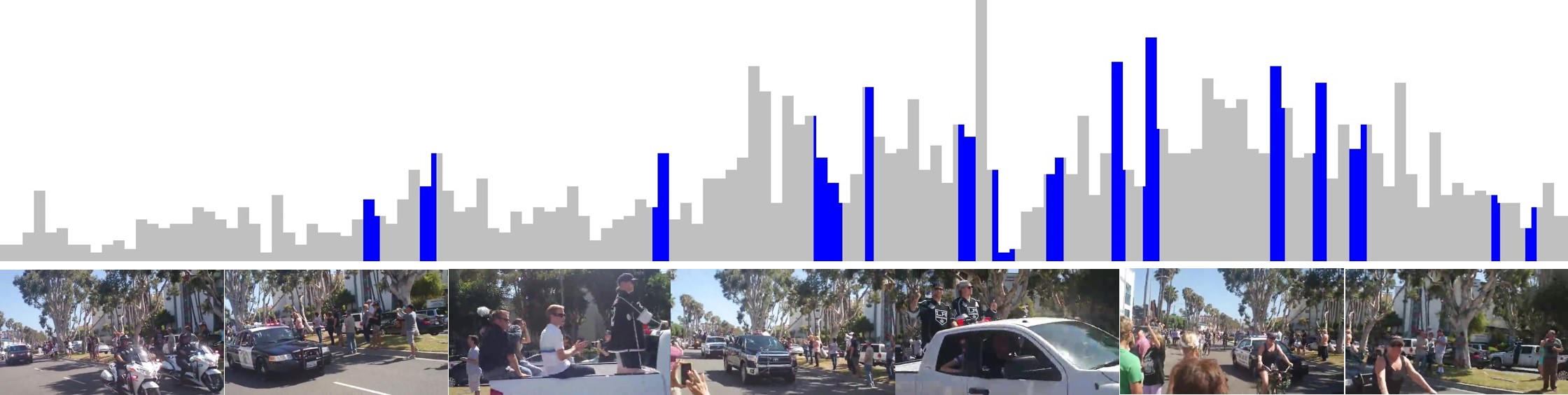}}\quad
  \subfloat[][video-46 (F-score: 59.1$\%$).]{\includegraphics[width=.49\textwidth]{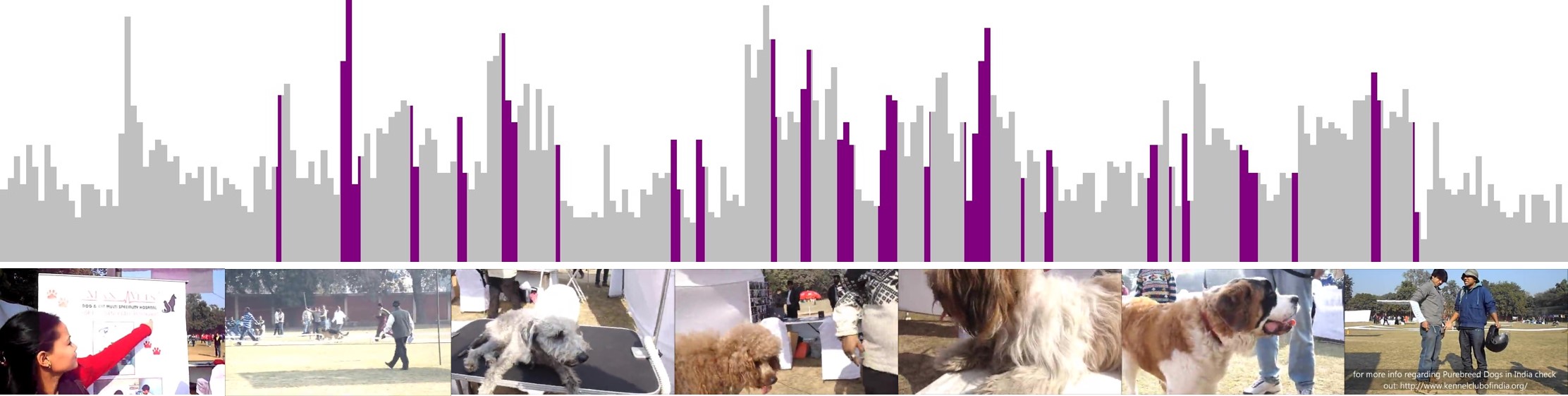}}
  \caption{The visualization of the generated summaries for four videos in TVSum. (a) video-18: "Poor Man's Meals: Spicy Sausage Sandwich". (b) video-22: "Charlotte Parkour". (c) video-28: "LA Kings Stanley Cup South Bay Parade". (d) video-46: "German Shepherd Dog Show".}
  \Description{The visualization of the generated summaries for four videos in TVSum. (a) video-18: "Poor Man's Meals: Spicy Sausage Sandwich". (b) video-22: "Charlotte Parkour". (c) video-28: "LA Kings Stanley Cup South Bay Parade". (d) video-46: "German Shepherd Dog Show".}
  \label{fig-four}
\end{figure*}

$M$ is not same for different videos according to the number of total frames $f^{total}$. $f^{clip}$ is the number of frames in a clip (16 by default). Empirically, the number of frames focused on by agents is around 15 percent of $f^{total}$, with the relationship of Eq.~\eqref{eq:relation}. Therefore, $N$ is also not same for different videos. In contrast experiments, $f^{clip}$ is the integer multiple of 16, and $N$ changes accordingly. For TVSum, if we take $f^{clip}$ to be $16$, then $f^{total}$ ranges from 2500 to 19406, $M$ ranges from 157 to 1213, and $N$ ranges from 25 to 182. For SumMe, if we take $f^{clip}$ to be $16$, then $f^{total}$ ranges from 950 to 9721, $M$ ranges from 60 to 608, and $N$ ranges from 9 to 92.

\begin{equation}
  N f^{clip} \leq 15\% f^{total}
  \label{eq:relation}
\end{equation}

Unlike other MARL-based methods, our $N$ is dynamic due to the characteristics of the video summarization task which requires 15-percent-duration summaries. We keep the optimization process stable by three means. Firstly, The agents are identical in network architecture and share the same parameters. Secondly, the reward values are within the same scope for different $N$. Thirdly, the optimization is carried out uniformly for all agents during an epoch. In this way, our method adapts to the video summarization task well with dynamic $N$ and achieves a good performance.

We set the action space as~\eqref{eq:action_space}, in which negative numbers denote moving left, zero means staying still, and positive numbers denote moving right (the unit is clip).

\begin{equation}
  u_t \in \{ -16, -8, -4, -2, -1, 0, +1, +2, +4, +8, +16 \}
  \label{eq:action_space}
\end{equation}

The scale factors $\alpha_1$, $\alpha_2$, $\alpha_3$ and $\alpha_4$ are set to $1$. $\gamma$ is set to $0.9$. We use Adam~\cite{kingma2014adam} with the initial learning rate of $\eta = 0.0001$. $\lambda$ is set to 1. The hidden state dimension is 512. The number of epochs for each video is $10$ ($K = 10$) and the number of episodes for a dataset is $5$. The maximum iteration in an epoch is $20$.

\begin{table}
  \caption{The evaluation results of CoSNet-US ($\%$) with different clip sizes on SumMe and TVSum.}
  \label{tab-clip}
  \begin{tabular}{c|c|c}
    \toprule
    Clip Size (frames) & SumMe & TVSum \\
    \midrule
    16 & 47.8 & 59.7 \\
    32 (16$\times$2) & 46.1 & 58.5 \\
    48 (16$\times$3) & 45.3 & 57.4 \\
    \bottomrule
  \end{tabular}
\end{table}

\subsection{Comparative Experiments}

The comparative experiments are conducted on different reward combinations and different clip sizes respectively.

\paragraph{\textbf{Reward Combinations}}

We conduct comparative experiments on different reward combinations. The complete reward definition is composed of supervised reward and unsupervised reward, each containing local part and global part, as mentioned in Sec.~\ref{sec:rd}. We experiment on seven reward combinations, complete reward (CoSNet-US), only unsupervised reward (CoSNet-U), only supervised reward (CoSNet-S), only local unsupervised reward (CoSNet-LU), only global unsupervised reward (CoSNet-GU), only local supervised reward (CoSNet-LS) and only global supervised reward (CoSNet-GS). By setting up the seven different reward combinations, we can draw the difference between unsupervised reward and supervised reward, and the difference between local parts and global parts. Fig.~\ref{fig-one} visualizes the generated summaries of video-17 in TVSum with four different reward combinations.

\begin{table}
  \caption{The evaluation results ($\%$) of CoSNet-U and other unsupervised methods on SumMe and TVSum.}
  \label{tab-unsupervised}
  \begin{tabular}{c|c|c}
    \toprule
    Method & SumMe & TVSum \\
    \midrule
    Video-MMR~\cite{li2010multi} & 26.6 & - \\
    Vsumm~\cite{de2011vsumm} & 33.7 & - \\
    Web image~\cite{khosla2013large} & - & 36.0 \\
    Co-archetypal~\cite{song2015tvsum} & - & 50.0 \\
    SUM-GAN$_{dpp}$~\cite{mahasseni2017unsupervised} & 39.1 & 51.7 \\
    DR-DSN~\cite{zhou2018deep} & 41.4 & 57.6 \\
    SUM-FCN$_{unsup}$~\cite{rochan2018video} & 41.5 & 52.7 \\
    Xiong {\itshape et al.}~\cite{xiong2019less} & - & 56.3 \\
    Rathore {\itshape et al.}~\cite{rathore2019generating} & 45.6 & 59.1 \\
    ACGAN~\cite{he2019unsupervised} & 46.0 & 58.5 \\
    \midrule
    \midrule
    CoSNet-U & $\textbf{46.4}$ & $\textbf{59.3}$ \\
    \bottomrule
  \end{tabular}
\end{table}

Tab.~\ref{tab-reward} shows the evaluation results of different reward combinations on SumMe and TVSum. Among the seven combinations, CoSNet-US with the complete reward outperforms other reward combinations which demonstrates that different kinds and different parts of the rewards together contribute to a good performance on both datasets.

Comparing CoSNet-S with CoSNet-U, we can find that supervised reward performs slightly better than unsupervised reward ($47.1\%$ vs. $46.4\%$ on SumMe and $59.4\%$ vs. $59.3\%$ on TVSum), which is the usual case in the contrast between unsupervised methods and supervised methods. Meanwhile, the difference in our methods is smaller than that of other methods (only $0.7\%$ on SumMe and $0.1\%$ on TVSum). The small difference justifies that the user summarization process modeling has less relationship with the subjective annotations than other methods.


We can also find that CoSNet-LS achieves higher F-score than CoSNet-GS ($46.3\%$ vs. $44.6\%$ on SumMe and $58.7\%$ vs. $57.3\%$ on TVSum) and CoSNet-LU achieves higher F-score than CoSNet-GU ($45.9\%$ vs. $43.5\%$ on SumMe and $58.4\%$ vs. $56.9\%$ on TVSum). The reason why methods with only local parts perform better than methods with only global parts is that the local parts in the rewards reflect local agent-level gains, which can benefit more from the multi-agent setting of {\itshape CoSNet} than the global parts.

Then, we can analyze this result further by comparing what the different reward parts have done respectively during the learning process. Specifically, for unsupervised reward, the local clip centrality ensures that each focus clip approximates and covers its neighboring clips, while the global feature difference only makes the summary diverse. Similarly, for supervised reward, the local change and local skip help to save more key frames at clip level, while the global change only help to save key frames at video level. Both global parts of the rewards can not benefit from the many agents which is the key characteristic of {\itshape CoSNet}.

\begin{table}
  \caption{The evaluation results ($\%$) of CoSNet-US and other supervised methods on SumMe and TVSum.}
  \label{tab-supervised}
  \begin{tabular}{c|c|c}
    \toprule
    Method & SumMe & TVSum \\
    \midrule
    Interestingness~\cite{gygli2014creating} & 39.4 & - \\
    Submodularity~\cite{gygli2015video} & 39.7 & - \\
    Summary transfer~\cite{zhang2016summary} & 40.9 & - \\
    Bi-LSTM~\cite{zhang2016video} & 37.6 & 54.2 \\
    DPP-LSTM~\cite{zhang2016video} & 38.6 & 54.7 \\
    SUM-GAN$_{sup}$~\cite{mahasseni2017unsupervised} & 41.7 & 56.3 \\
    DR-DSN$_{sup}$~\cite{zhou2018deep} & 42.1 & 58.1 \\
    MAVS~\cite{feng2018extractive} & 43.1 & $\textbf{67.5}$ \\
    DySeqDPP~\cite{li2018local} & 44.3 & 58.4 \\
    ACGAN$_{sup}$~\cite{he2019unsupervised} & 47.2 & 59.4 \\
    SUM-FCN~\cite{rochan2018video} & 47.5 & 56.8 \\
    SUM-DeepLab~\cite{rochan2018video} & 48.8 & 58.4 \\
    SMN~\cite{wang2019stacked} & $\textbf{58.3}$ & 64.5 \\
    \midrule
    \midrule
    CoSNet-US & 47.8 & 59.7 \\
    \bottomrule
  \end{tabular}
\end{table}

\paragraph{\textbf{Clip Sizes}}

We also conduct comparative experiments on different clip sizes. In the {\itshape Implementation Details} part of Sec.~\ref{sec:es}, we introduce the relationship of $N$, $f^{clip}$ and $f^{total}$ with Eq.~\eqref{eq:relation}. The default clip size $f^{clip}$ is 16 frames. For comparison, we further expand $f^{clip}$ to 32 frames ($2\times16$) and 48 frames ($3\times16$), and $N$ changes accordingly. The C3D features is extracted every 16 frames. Therefore, the input feature vector for larger $f^{clip}$ is the average of two or three C3D feature vectors.

Tab.~\ref{tab-clip} shows the evaluation results of CoSNet-US with different clip sizes on SumMe and TVSum. The F-score decreases with the increase of clip size, which indicates that the clip size of 16 is moderate for the video summarization task. Smaller clips can be more accurate in granularity than larger clips when making comparisons and selections.


\begin{table}
  \caption{The evaluation results ($\%$) of CoSNet-U and other unsupervised methods on SumMe and TVSum under Augmented and Transfer settings.}
  \label{tab-setting}
  \begin{tabular}{c|c|c|c|c}
    \toprule
    \multirow{2}{*}{Method} & \multicolumn{2}{c}{SumMe} & \multicolumn{2}{|c}{TVSum} \\
    \cmidrule{2-5}
    & A & T & A & T \\
    \midrule
    SUM-GAN$_{dpp}$~\cite{mahasseni2017unsupervised} & 43.4 & - & 59.5 & - \\
    DR-DSN~\cite{zhou2018deep} & 42.8 & 42.4 & 58.4 & 57.8 \\
    ACGAN~\cite{he2019unsupervised} & 47.0 & 44.5 & 58.9 & 57.8 \\
    \midrule
    \midrule
    CoSNet-U & 47.1 & 45.2 & 59.8 & 58.1 \\
    \bottomrule
  \end{tabular}
\end{table}

\begin{table}
  \caption{The evaluation results ($\%$) of CoSNet-U on YouTube, OVP, SumMe and TVSum under one-to-one Transfer setting.}
  \label{tab-one2one}
  \begin{tabular}{c|c|c|c|c}
    \toprule
    Test$\setminus$Training & YouTube & OVP & SumMe & TVSum \\
    \midrule
    SumMe & 42.8 & 43.5 & - & 44.1 \\
    \midrule
    TVSum & 56.5 & 57.4 & 56.6 & - \\
    \bottomrule
  \end{tabular}
\end{table}

\subsection{Method Comparison}

We compare {\itshape CoSNet} with other methods. For comparison with unsupervised methods, we use CoSNet-U with only unsupervised reward. For comparison with supervised methods, we use CoSNet-US with the complete reward. We also make comparisons under Augmented and Transfer settings. To ensure fair comparison, the numerical results are mostly from \cite{he2019unsupervised} and partly from other state-of-the-art methods not included in \cite{he2019unsupervised}.

\paragraph{\textbf{Comparison with Unsupervised Methods}}

Tab.~\ref{tab-unsupervised} shows the evaluation results of CoSNet-U and other unsupervised methods on SumMe and TVSum. CoSNet-U outperforms state-of-the-art unsupervised methods on both datasets. Specifically, CoSNet-U performs $0.4\%$ better than ACGAN on SumMe and $0.2\%$ better than the method by Rathore {\itshape et al.} \cite{rathore2019generating} on TVSum. The performance with unsupervised reward demonstrates that CoSNet-U solves the subjectivity problem to some extent.

\paragraph{\textbf{Comparison with Supervised Methods}}

\begin{figure*}[t]
  \centering
  \subfloat[][video-2 (F-score: 65.1$\%$).]{\includegraphics[width=.49\textwidth]{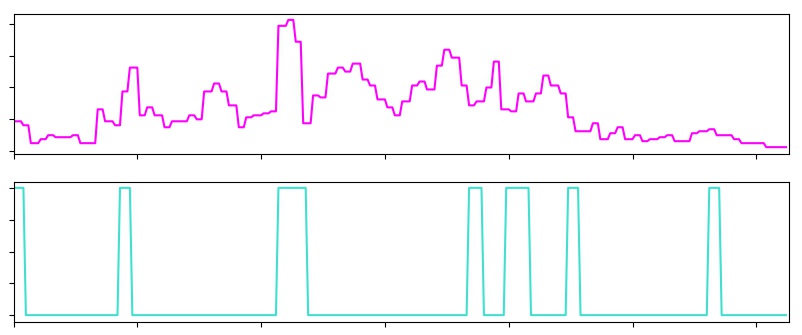}}\quad
  \subfloat[][video-36 (F-score: 67.1$\%$).]{\includegraphics[width=.49\textwidth]{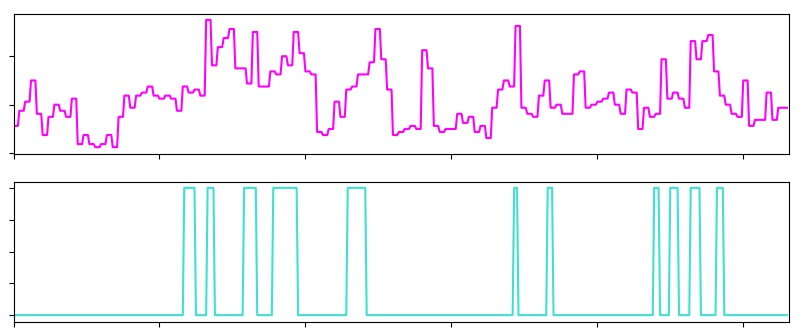}}\\
  \subfloat[][video-39 (F-score: 66.7$\%$).]{\includegraphics[width=.49\textwidth]{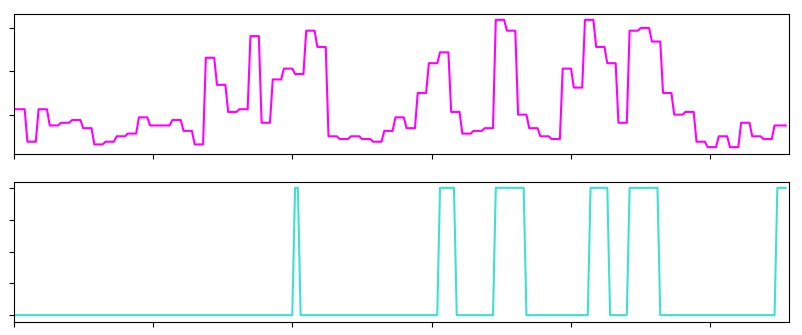}}\quad
  \subfloat[][video-41 (F-score: 84.6$\%$).]{\includegraphics[width=.49\textwidth]{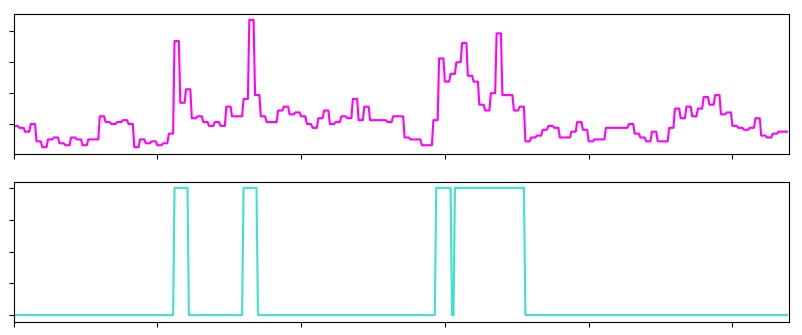}}
  \caption{The contrast between the ground truth importance score distributions of the given summaries and the binary score distributions of the generated summaries for video-2, video-36, video-39 and video-41 in TVSum.}
  \Description{The contrast between the ground truth importance score distributions of the given summaries and the binary score distributions of the generated summaries for video-2, video-36, video-39 and video-41 in TVSum.}
  \label{fig-distribution}
\end{figure*}

Tab.~\ref{tab-supervised} shows the evaluation results of CoSNet-US and other supervised methods on SumMe and TVSum. CoSNet-US outperforms most supervised methods on both datasets. Meanwhile, the F-score of CoSNet-US is lower than SUM-DeepLab (SumMe), MAVS (TVSum) and SMN (both). Besides, the performance of MAVS on SumMe is inferior to CoSNet-US, and the performance of SUM-DeepLab on TVSum is also inferior to CoSNet-US. The reason is that the supervised reward brings some promotion with the annotation information, but also brings the inherent subjectivity in the annotations.

\paragraph{\textbf{Comparison under Augmented and Transfer Settings}}

Tab.~\ref{tab-setting} shows the evaluation results of CoSNet-U and other unsupervised methods on SumMe and TVSum under Augmented and Transfer settings. CoSNet-U outperforms other unsupervised methods on both datasets under both settings.

Tab.~\ref{tab-one2one} shows the evaluation results of CoSNet-U on YouTube, OVP, SumMe and TVSum under one-to-one Transfer setting. The training sets include all the four datasets, while the test sets only include SumMe and TVSum. With only one dataset for training, CoSNet-U performs worse than with three datasets for training. The reason is that the number of videos in the training set with three datasets is larger than with only one dataset, which also applies to the difference among training sets with only one dataset (TVSum-50 $=$ OVP-50 $>$ YouTube-39 $>$ SumMe-25).

\subsection{Qualitative Analysis}

We analyze {\itshape CoSNet} qualitatively with visual illustrations in this subsection. Fig.~\ref{fig-one} and Fig.~\ref{fig-four} are the visualization for the generated summaries of five videos in TVSum. The heights of all the bars (grey or colored) represent the ground truth importance scores of the videos, while the colored bars represent the selected segments. We can find that most bars with relatively higher ground truth importance scores are selected by different varieties of {\itshape CoSNet}.

In Fig.~\ref{fig-one}, there is a comparison among four different methods with different reward combinations on video-17 in TVSum. The four methods (CoSNet-US, CoSNet-U, CoSNet-S and CoSNet-LU) perform likely with slight difference in the selection of segments.

Fig.~\ref{fig-distribution} shows the contrast between the ground truth importance score distributions of the given summaries and the binary score distributions of the generated summaries. With {\itshape CoSNet}, the summarization results are in the form of key clips, which can not be directly transformed into continuous scores ($[0,1]$). Therefore, we make comparisons between two different forms of scores, which can still demonstrate the quality of the generated summaries intuitively. We can find that the high scores of the given ground truth summaries and the "one" scores of the generated summaries appear in similar segments.

\section{Conclusion}

In this paper, we propose Comparison-Selection Network ({\itshape CoSNet}) based on multi-agent reinforcement learning for video summarization. In order to solve the subjectivity problem and inspired by the general user behaviours, we formulate the user summarization process as multiple sequential decision-making processes. Each process is modeled by a comparison network for feature extraction and a selection network for decision making. The unsupervised reward and supervised reward, each with local and global parts, together contribute to the advancement of the policy. Experiments on two benchmark datasets show that {\itshape CoSNet} outperforms state-of-the-art unsupervised methods and most supervised methods.


\bibliographystyle{ACM-Reference-Format}
\bibliography{sample-sigconf}


\end{document}